\documentclass[%
 reprint,
 amsmath,amssymb,
 aps,
]{revtex4-2}
\usepackage{hyperref}
\usepackage{graphicx}
\usepackage{dcolumn}
\usepackage{bm}

\begin{document}

\preprint{APS/123-QED}

\title{Model-Free Prediction of Chaotic Systems Using High Efficient Next-generation Reservoir Computing}
\thanks{Project supported by the China Postdoctoral Science Foundation (No. 2019T120447).}%

\author{Zhuo~Liu}
\author{Leisheng~Jin}%
\email{jinls@njupt.edu.cnl}
\affiliation{College of Electronic and Optical Engineering, \\Nanjing University of Posts and Telecommunications, Nanjing 210023, China}

\date{\today}

\begin{abstract}
To predict the future evolution of dynamical systems purely from observations of the past data is of great potential application. In this work, a new formulated paradigm of reservoir computing is proposed for achieving model-free predication for both low-dimensional and very large spatiotemporal chaotic systems. Compared with traditional reservoir computing models, it is more efficient in terms of predication length, training data set required and computational expense. By taking the Lorenz and Kuramoto-Sivashinsky equations as two classical examples of dynamical systems, numerical simulations are conducted, and the results show our model excels at predication tasks than the latest reservoir computing methods.
\end{abstract}

\maketitle

Reservoir computing (RC), as a machine learning based technique, has been attracting widely interests due its outstanding performance in not only recognition tasks but also multivariate time series prediction of dynamical systems \cite{sc1} \cite{sc2} \cite{sc3} \cite{WCY}. For the latter, various improved RC models in terms of structure modification have been proposed. These include double-reservoir in parallel RC (DRESN) \cite{DRESN},  broad-ESN \cite{BESN}, hierarchical delay-memory echo state network (HDESN) \cite{HDESN}, integer echo state networks (intESN) \cite{intESN} etc. Particularly, L. Appeltant et al. introduced an architecture of RC that only used a single dynamical node with delay feedback--time-delay reservoir (TDR)\cite{LAN}. Very recently, the so-called next generation reservoir computing (NG-RC) which is characterized by using nonlinear vector autoregression (NVAR) to replace traditional reservoir is newly developed \cite{NVAR} \cite{NG-RC}. The NG-RC works excellently at reservoir computing benchmark tasks such as forecasting, reproducing and inferring unseen data/behavior of a dynamical system. However, it still stays at the phase for studying low dimensional chaotic systems, and if applied the NG-RC for very large spatiotemporal chaotic systems (VLSCS), a predictable problem arising is that the computing expense become very high, as the VLSCS brings much more terms into the feature vector construction. Therefore, advanced techniques require even smaller training data sets and minimal computing resources, together with holding the functionality of dealing with a wide scope of dynamical systems, particularly the VLSCS, are still high desired.

\begin{figure}[htb]
\centering
\includegraphics[width=0.99\linewidth]{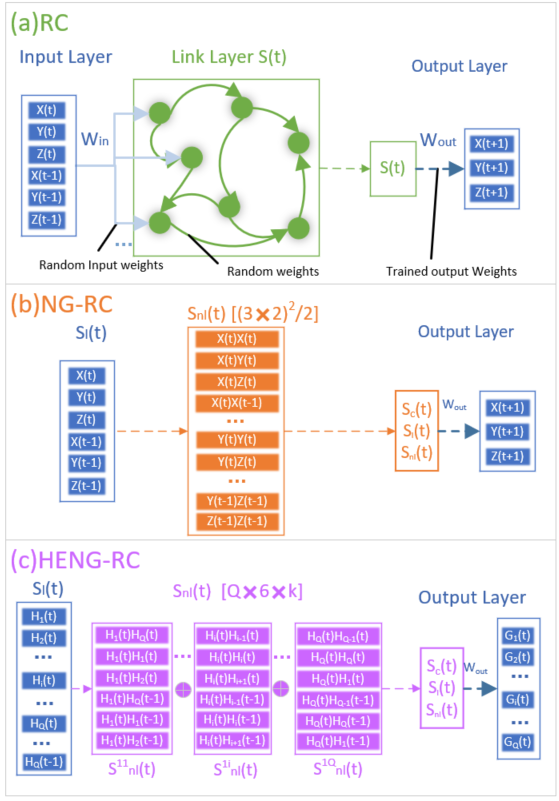}
\caption{\label{fig:models} (a) The structure of the traditional reservoir computing; (b) The original NG-RC; (c)The proposed HENG-RC.}
\end{figure}

In this work, we formulate a high efficient NG-RC (HENG-RC) that only takes the neighbour states coupling into account for constructing the feature vector (as detailed in Fig. \ref{fig:models}c), which can greatly reduce the nonlinear terms demanded as in the original NG-RC and increase the computational speed. The HENG-RC not only works effectively for low-dimensional chaotic systems but also for VLSCS. The Lorenz system is first used for verifying the effectiveness of the proposed model, with more satisfactory results yield. More importantly, we extend the technique into the predication of VLSCS. By using Kuramoto-Sivashinsky equation as an illustrative model, it is proven that the proposed HENG-RC can achieve an excellent prediction results with a much less computational sources than traditional RC-based method. The work offers an effective RC-based technique for model-free predication of both the low-dimensional systems and VLSCS. 
We briefly describe the working principle of the traditional reservoir computing as shown in Fig.\ref{fig:models}a. It is generally composed of three parts: an input layer, a reservoir and an output layer. The input layer feeds the input signal (vector) via a weight matrix $W_{in}$ to the reservoir composed of $N$ interconnected nodes. The strength of connections among these nodes is randomly generated and keeps fixed. The output vector can be derived through weight matrix $W_{out}$ which couples the states of reservoir nodes and output layer. 
Assuming a three-dimensional vector $u(t)$ composed of $X(t), Y(t)$ and $Z(t)$ with time $t$ in discrete as input signal, the dynamics of the nodes in the reservoir can be described by \cite{dy}:
\begin{gather}\label{equ:rc_s}
    S(t+1) = (1-\gamma) S(t) + \gamma f[S(t) A +W_{in}u(t)+b]
\end{gather}
where $S(t)$ is a $N$-dimensional vector representing the state of $N$ nodes at time $t$. The $f$ is an activation function, $\gamma$ is the decay rate of the nodes, and $b$ is a node bias vector. The output vector is calculated by multiplying the $S(t)$ and the output weight matrix $W_{out}$, i.e.,:
\begin{gather}
\label{equ:output}
    Y'=W_{out}\times S
\end{gather}
There are two phases: training and predicating, included in the operation of traditional reservoir computing. In the training phase, the output weight matrix can be adjusted via a regularized linear least-squares optimization procedure, which is expressed as \cite{reg2}:
\begin{gather}
\label{equ:regression}
    W_{out}= Y S^T (S S^T+ \lambda I)^{-1}
\end{gather}
where $Y$ is the output we desired, $I$ is an identity matrix and $\lambda$ is ridge parameter to prevent over-fitting. After trained, the $W_{out}$ is fixed, and one can let the output vector $Y'$ feed back into input layer, the reservoir computer can automatically run itself for generating future states, i.e., the predicating phase.

Vastly different from the traditional reservoir computing, the next generation reservoir computing (NG-RC) \cite{NG-RC}, as shown in Fig.\ref{fig:models}b, uses nonlinear vector autoregression (NVAR) to replace the reservoir. The NG-RC creates a so-called feature vector using the input data at states at time $t$ and $k$ delay time states. The feature vector is consisted of two parts: linear and nonlinear. The linear part is composed of the input vector from time $t$ to $t-k$ ($k=1, 2, ...$), and the nonlinear part is composed of the outer product of the linear vector. The mathematical process for feature construction can be described by:
\begin{gather}
    S= S_c \oplus S_{l} \oplus S_{nl} \\
    S^p_{nl}= S_{l} \otimes S_{l} ... \otimes S_{l} \\
    S^k_l(t)= u(t) \oplus u(t-1) ... \oplus u(t-k) \notag
\end{gather}
where $\oplus$ represents the vector concatenation operation, $\otimes$ represents the outer product operation and $p$ is the number of outer product operation. $S_c$ is a constant vector, $S^k_l$ and $S_{nl}$ represent the linear and nonlinear state vector, respectively. The $k$ denotes the number of time-delay states. 



The NG-RC can be trained as same as the traditional reservoir computing. Compared with traditional reservoir computing, the NG-RC establishes direct connection of input data itself rather than uses an already linked nodes (reservoir), reducing the unnecessary connections to make it more concise and easy to be implemented for low-dimensional signal. However, the NG-RC would become inefficient for calculating the VLSCS, as there are much more linear and nonlinear terms brought by VLSCS to create the feature vector. 

Our proposed HENG-RC is given in Fig.\ref{fig:models}c, which has a distinctive feature vector comparing with the NG-RC. The most difference from the original NG-RC is that the nonlinear part of feature vector are composed of only the products of neighboring linear states rather than outer product of all linear states, where the coupling between linear and nonlinear parts is newly defined, which can be mathematically expressed by Equ. \ref{equ:HENG-RCn}:

\begin{gather}
\label{equ:HENG-RCn}
    S_{nl}= S^1_{nl} \oplus S^{2}_{nl} ... \oplus S^{k}_{nl} 
\end{gather}
\begin{gather}
\label{equ:HENG-RCn_nl}
    S^{j}_{nl} = S^{1i}_{nl} ... \oplus S^{ji}_{nl} \oplus... S^{Qi}_{nl}\\
    S^{ji}_{nl} = H_{i}(t-j) \times H_{i-1}(t-j) \notag \\
    \oplus H_i(t-j) \times H_{i}(t-j)  \notag \\
    \oplus H_i(t-j) \times H_{i+1}(t-j) \notag \\
    \oplus H_{i}(t-j) \times H_{i-1}(t-j-1) \notag \\
    \oplus H_i(t-j) \times H_{i}(t-j-1)  \notag \\
    \oplus H_i(t-j) \times H_{i+1}(t-j-1) \notag 
\end{gather}

where $H(t)$ is a $Q$-dimensional input vector in general. Assuming there are $k$ time-delay states considered, the overall nonlinear part $S_{nl}$, Eq. \ref{equ:HENG-RCn}, is divided into $k$ individual parts: $S^j_{nl}$ (j=1,2...k). Each of $S^j_{nl}$ is further divided to $Q$ sub-vectors $S^{ji}_{nl}$, which corresponds to dimension of the input vector. The $S^{ji}_{nl}$ is composed of the products between $H_{i}(t-j)$ and the three neighboring dimensional states ($i-1, i, i+1$) in input vector at time $t-j$ and time $t-j-1$, i.e., : $H_{i-1}(t-j), H_{i}(t-j), H_{i+1}(t-j), H_{i-1}(t-j-1), H_{i}(t-j-1), H_{i+1}(t-j-1)$. 

Taking the three-dimensional Lorenz system as an example, the Equ.\ref{equ:HENG-RCn} can be expressed as:
\begin{gather}
\label{equ:HENG-RC3}
    S_{nl}=  X(t) \times X(t) \oplus  X(t) \times Y(t) \oplus X(t) \times Z(t) \notag \\
    X(t) \times X(t-1) \oplus X(t) \times Y(t-1) \oplus X(t) \times Z(t-1) \notag
\end{gather}
It is concluded that the terms in the nonlinear part for constructing the feature vector is $Q\times 6\times k$, while the number in original NG-RC is $(Q\times(k+1))^2/2$. The computational expense is therefore lower. The training for our HENG-RC is as same as traditional RC and NG-RC.
 


Numerically, we first use the proposed model to predict the Lorenz system. The Lorenz system was first proposed in 1963 by Edward N. Lorenz \cite{Lorenz1963}, and it becomes one of the most famous nonlinear model for studding chaos. The model of Lorenz system is expressed as:
\begin{gather}
\dot{x} = \sigma (y-x) \notag \\
\dot{y} = \gamma x-y-xz \notag\\
\label{Equ:Lorenz}\dot{z} = x y-\beta z
\end{gather}
where $\sigma=10$, $\gamma=28$ and $\beta=8/3$, and in such a parameter setting the system works in chaotic state. 

\begin{figure}[htb]
\centering
\includegraphics[width=0.99\linewidth]{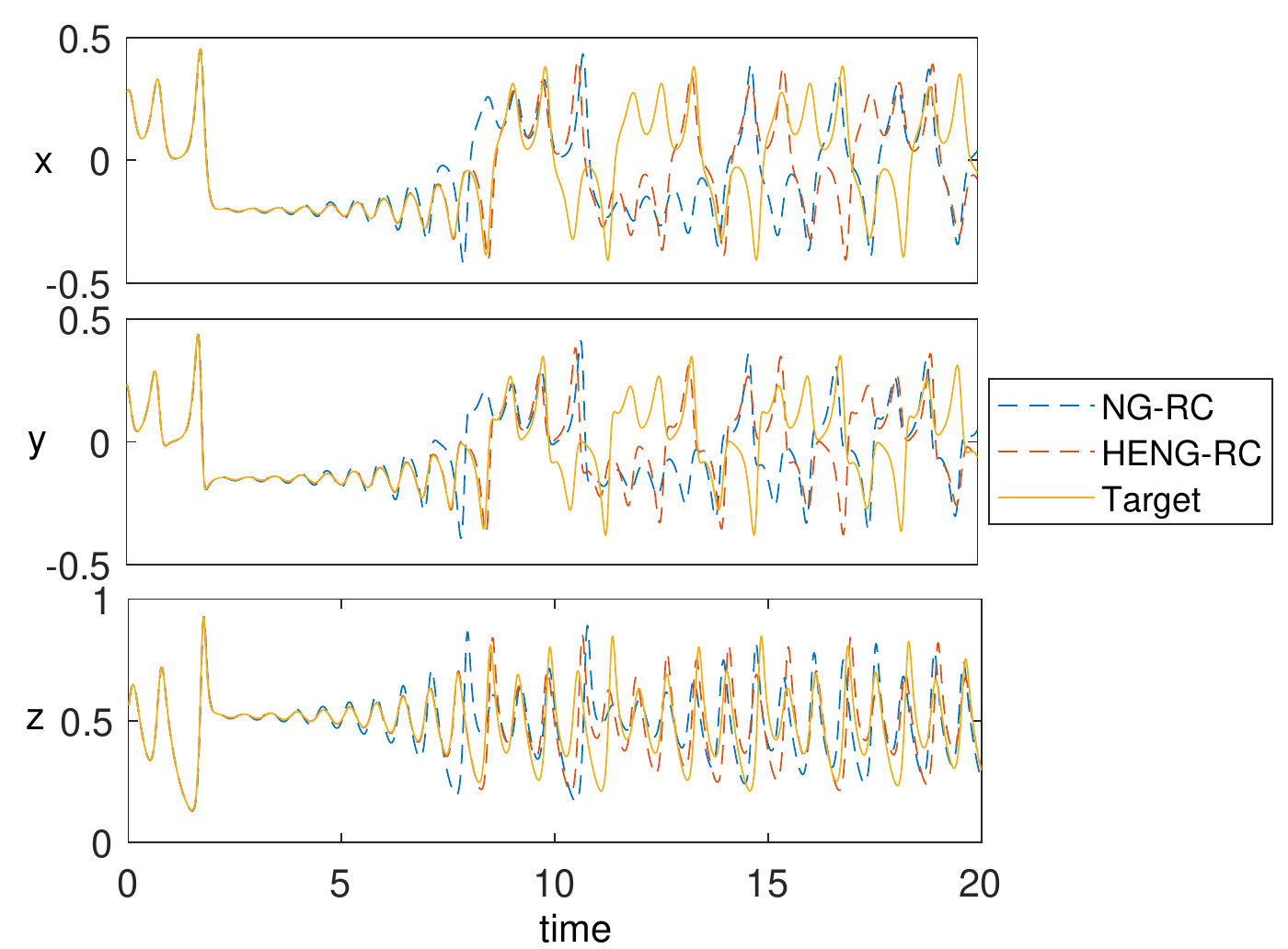}
\caption{\label{fig:LT4} The prediction of Lorenz system using NG-RC and HENG-RC with a comment training data set having 400 time steps.}
\end{figure}

\begin{figure}[htb]
\centering
\includegraphics[width=0.99\linewidth]{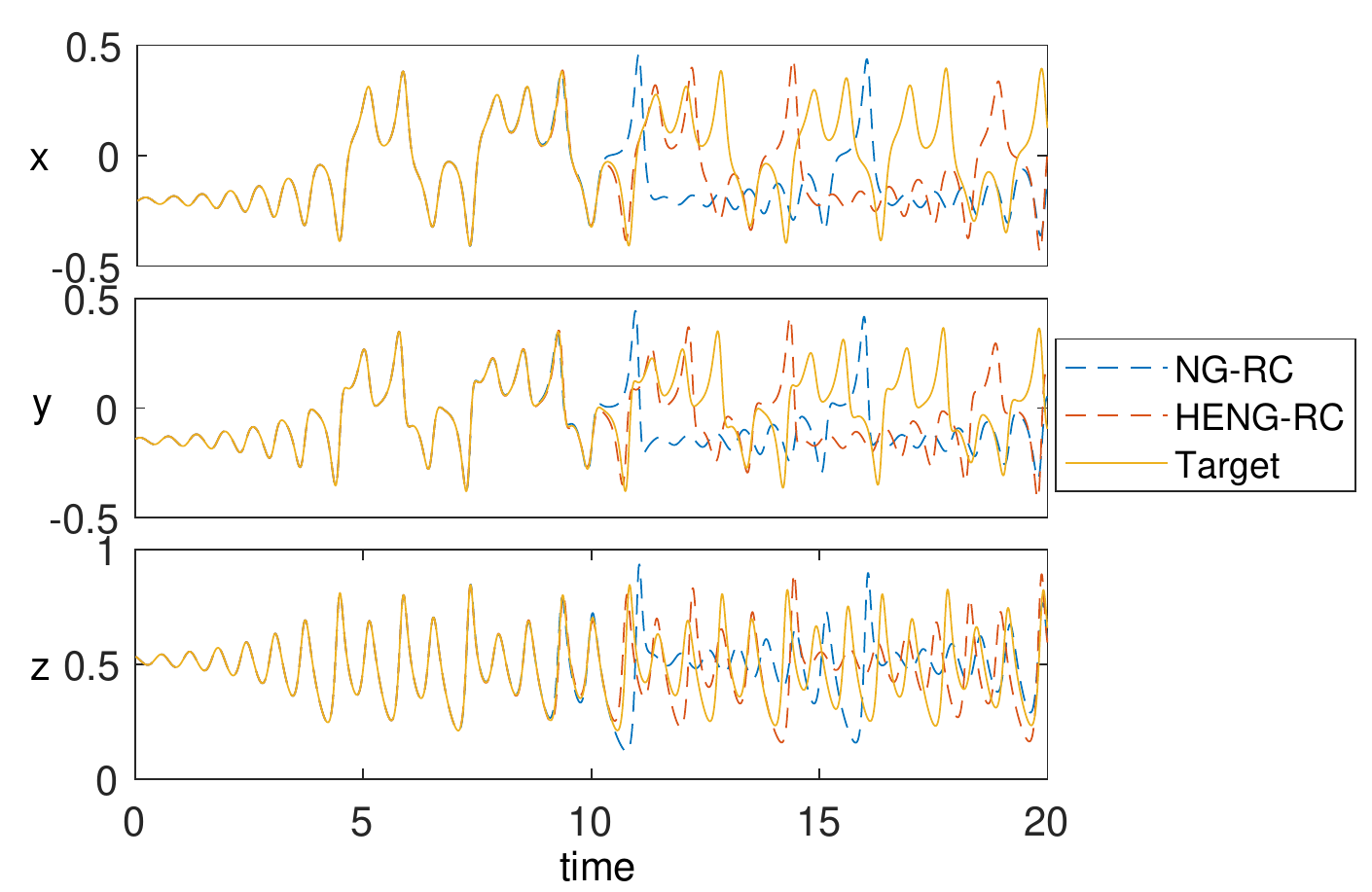}
\caption{\label{fig:LT8}  The prediction of Lorenz system using NG-RC and HENG-RC with a comment training data set having 800 time steps.}
\end{figure}

\begin{center}
{\footnotesize{\bf Table 1.} Comparison of Prediction performance using NG-RC and RC.\\
\vspace{2mm}
\begin{tabular}{cccc}
\hline
{Model} & {Training$^{\rm a}$} & {Prediction$^{\rm a}$} & {States} \\\hline
RC & 2 & 3.27 & 28 \\
RC & 4 & 6.54 & 28 \\
RC & 8 & 10.22 & 28 \\
HENG-RC & 2 & 6.37 & 12 \\
HENG-RC & 4 & 9.10 & 12 \\
HENG-RC & 8 & 10.54 & 12 \\
\hline
$^{\rm a}$ time steps $\times \Delta t $\\
\end{tabular}}
\end{center}
The Lorenz system is solved by fourth-order Runge-Kutta method with step $\Delta t=0.01s$. Based on HENG-RC and NG-RC, respectively, we calculate the predication length of all three-dimensional states ($x, y, z$) using a common training data sets with $400$ time steps. As shown in Fig.\ref{fig:LT4}, the HENG-RC excels at the predication task with the accurate predication length reaching 1000 time steps ahead while the NG-RC is 700. To further prove the point, we enlarge the training data set to 800, and it is shown that our HENG-RC can predict about 1200 time steps, better then the result about 1100 using NG-RC. It is worth to emphasize that our HENG-RC only uses 6 nonlinear terms for constructing the feature vector while there are 18 in original NG-RC. We also conduct the study of comparison with traditional RC for the same predicting task using Lorenz system. The results are summarized in Table 1. It is shown that based on a same training data sets, the HENG-RC yields longer predicating length but with using less internal dynamical states.

Furthermore, the standard Kuramoto-Sivashinsky (KS) equation \cite{KS} is used as a VLSCS for verifing our HENG-RC, which is given by:
\begin{gather}
    y_t= -y y_x - y_{xx} - y_{xxxx}
    \label{equ:ks}
\end{gather}
Where the $y(x,t)$ is a scalar field. The scalar field is periodic in the interval $[0, L)$. $L$ can be seen as the scalar parameter in spatial dimension. The Equ.\ref{equ:ks} can be solved numerically for generating a data set $H(t)$. Here, the $H(t)$ is derived with a time step of $0.25s$ on a grid of $Q$ equidistant points, and thus the $H(t)$ can be considered as a $Q$-dimensional sequence. $Q$ is an integer larger than $L$.

\begin{figure}[htb]
\centering
\includegraphics[width=0.99\linewidth]{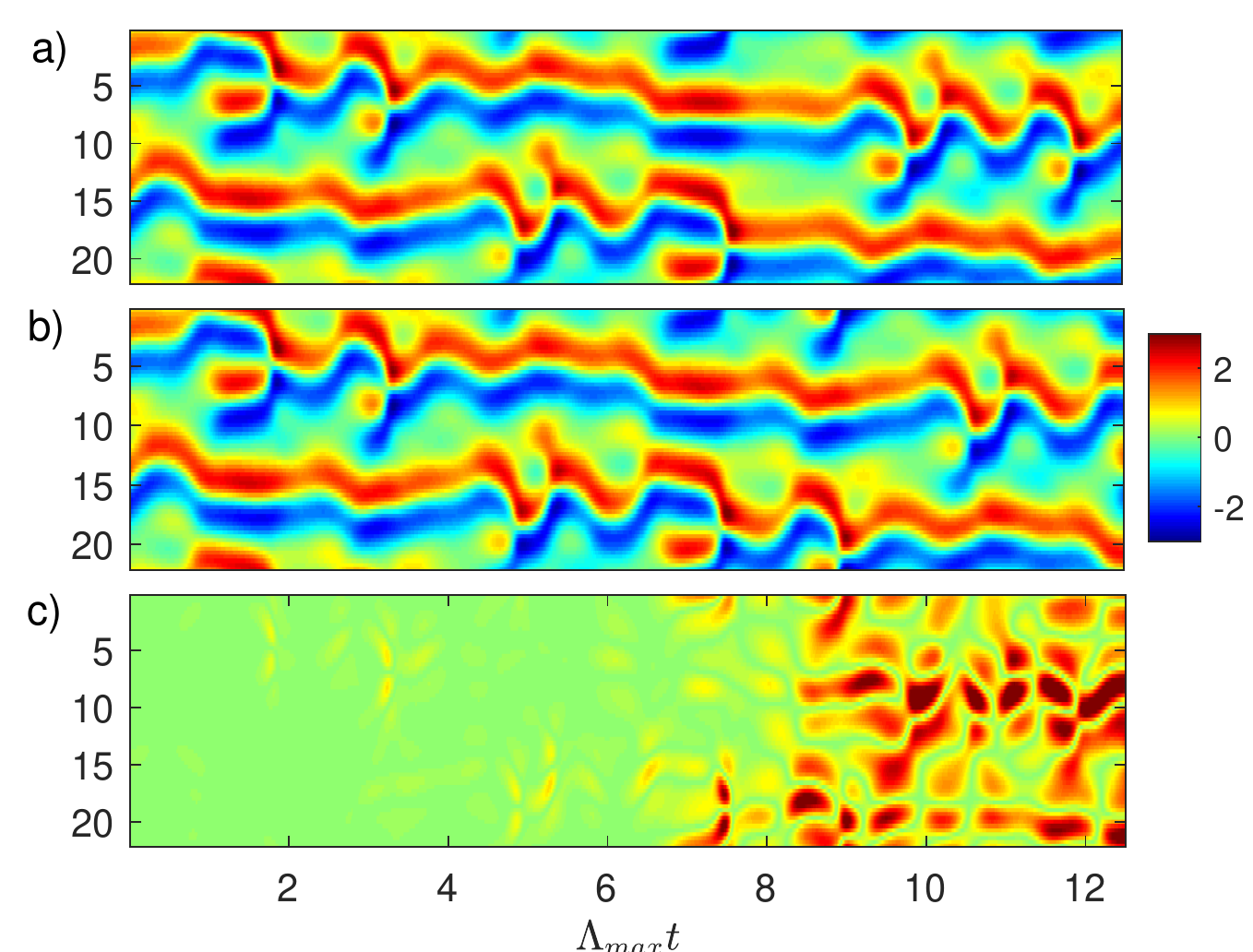}
\caption{\label{fig:L22} a) Spatiotemporal chaos generated by KS equation ($L=22, Q=64$); b) Prediction results of KS using the proposed HENG-RC; c) Difference between predication results and and the actual data.}
\end{figure}

First, we study the case based on HENG-RC when the $L$ is relatively small ($L=22$). A $H(t)$ is generated in this case, and the $Q$ is taken to be 64. Part of $H(t)$ is used for training the HENG-RC, and the rest data is used for verifying the predication accuracy. The result is present in Fig.\ref{fig:L22}. It is seen that the HENG-RC can predict accurately $7$ largest lyapunov time, confirming the effectiveness of our HENG-RC for the model-free prediction of VLSCS. 

\begin{figure}[htb]
\centering
\includegraphics[width=0.99\linewidth]{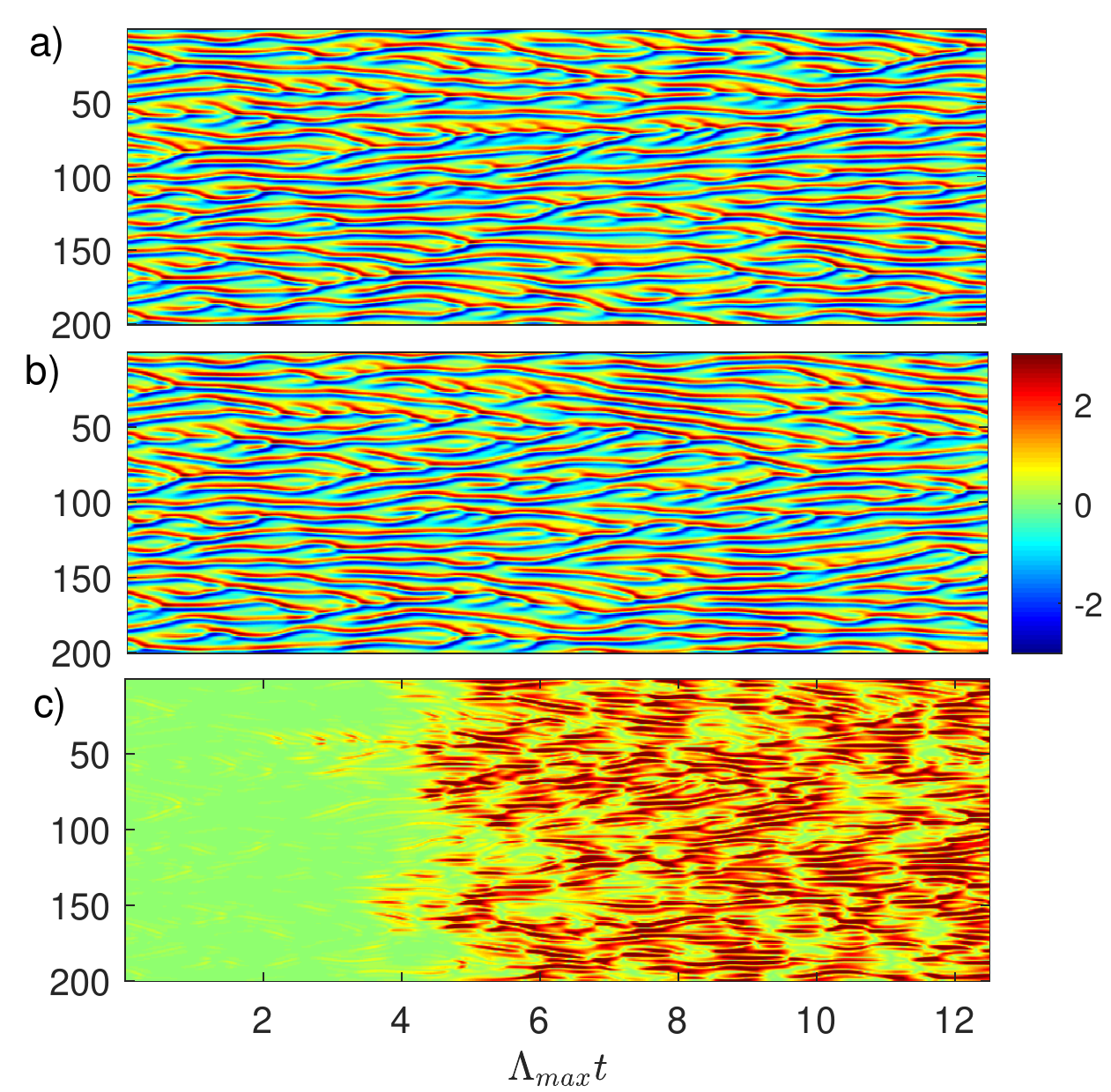}
\caption{\label{fig:L200} a) Spatiotemporal chaos generated by KS equation ($L=200, Q=64$); b) Prediction results of KS using the proposed HENG-RC; c) Difference between predication results and and the actual data.}
\end{figure}

Now, we increase the $L$ to 200 evolving into a large spatiotemporal dynamical systems. The predication task in this case is studied, and the result is given in Fig.\ref{fig:L200}, where it is shown that our HENG-RC can also yield a good result with the predication length reaching about $4$ largest lyapunov time.

To highlight the high efficiency of HENG-RC in dealing with VLSCS, the comparison with other RC-based techniques including RC, RC in parallel and NG-RC, is conducted and present. The results are summarized in Table. 2. There are two features can be concluded in the results that for the prediction task, i.e., the same $L$, (1): the HENG-RC uses the least states. For example, when $L=22$ the RC and NG-RC uses 3968 and 8384 states while the HENG-RC uses only 904; (2): The computational time cost is the lowest. For example, for $L=22$ the RC and NG-RC takes 132.24s and 290.98s while the HENG-RC is 10.438s. Therefore, the efficiency both in time cost and computational resource are higher. In addition, for larger $L$, e.g., $L=400$, our HENG-RC can perform excellently using almost the computational resource such that when using NG-RC to deal the case of $L=22$.

\begin{center}
{\footnotesize{\bf Table 2.} The comparison of computational efficiency using different RC based techniques. \label{tab:parallel parameters}\\
\vspace{2mm}
\begin{tabular}{ccccc}
\hline
{Model} & {L} & {Q} & {States} & {Time cost$^{\rm a}$} \\\hline
RC{\cite{PRL_Large}} & 22 & 64 & 3968 & 132.24\\
Parallel RC{\cite{PRL_Large}} & 200 & 512 & 320000 & Unknown\\
NG-RC & 22 & 64 & 8384 & 260.98\\
HENG-RC & 22 & 64 & 904 & 10.438\\
HENG-RC & 200 & 256 & 3592 & 167.11\\
HENG-RC & 400 & 512 & 7176 & 349.25\\
\hline
$^{\rm a}$CPU times calculated by Matlab.\\
\end{tabular}}
\end{center}
To explain why our HENG-RC works more efficient with less computational resources, the reason behind can be attributed to that the outer product operation for constructing the feature vector in NG-RC creates redundant states and affects data fitting. Specifically, for the low-dimensional chaotic system, it is found that the $\lambda$, an important parameter used for fitting in the training process, needs to be chosen larger than HENG-RC, which means the outer product in NG-RC may bring many unhelpful terms. For the VLSCS, based on HENG-RC, the over-fitting error caused by accidental spatial closeness in VLSCS can be reduced. At the same time, the construction of feature vector is limited to the adjacent state in time and space, which can avoid the accidental instability that may be occurred in NG-RC. So the HENG-RC can improve the robustness and speed obviously compared with the NG-RC. 

In conclusion, a new formulation of NG-RC, i,e., the HENG-RC, is present. It is characterized by its high efficiency in dealing with both the low-dimensional and very large spatiotemporal chaotic systems. Based on limited observational data in the past of dynamical systems, the model-free accurate predication can be achieved. The work lays a foundation for using next generation reservoir computing to predicate various types of chaotic systems with a high efficiency.


\addcontentsline{toc}{chapter}{References}
\begin{thebibliography}{99}\footnotesize
\itemsep=-3pt plus.2pt minus.2pt   
\bibitem{sc1} Azarakhsh J, Kris D,  Wesley D N, and Jean-Pierre M \href{https://www.sciencedirect.com/science/article/pii/S0925231217314145}{2018 \emph{Neurocomputing} \textbf{277} 237}

\bibitem{sc2} Long J Y, Zhang S H and Li C  \href{https://ieeexplore.ieee.org/document/8827669}{2018 \emph{IEEE Trans Industr Inform} \textbf{16} 4928}

\bibitem{sc3} Bo Y C, Wang P and Zhang X  \href{https://www.sciencedirect.com/science/article/pii/S1568494620304695}{2020 \emph{APPL SOFT COMPUT} \textbf{95} 106530}

\bibitem{WCY}Weng T F, Cao X X, and Yang H J \href{http://cpb.iphy.ac.cn/EN/10.1088/1674-1056/abd9b3}{2021 \emph{Chin. Phys. B} \textbf{30} 060506}

\bibitem{DRESN}Zhong S S, Xie X L, Lin L, and Wang F \href{https://www.sciencedirect.com/science/article/pii/S0925231217301273}{2017 \emph{Neurocomputing} \textbf{238} 191}

\bibitem{BESN}Yao X S and Wang Z S \href{https://www.sciencedirect.com/science/article/pii/S0925231217301273}{2019 \emph{J Franklin Inst} \textbf{356} 4888}

\bibitem{HDESN}Na X D, Ren W J, and Xu X H \href{https://www.sciencedirect.com/science/article/pii/S0952197621000762}{2021 \emph{Eng Appl Artif Intell} \textbf{102} 104229}

\bibitem{intESN}Kleyko D, Paxon Frady E, Kheffache M, and Osipov E \href{https://pubmed.ncbi.nlm.nih.gov/33351770/}{2020 \emph{IEEE T NEUR NET LEAR}}

\bibitem{LAN}L. Appeltant, M.C. Soriano, G. Van der Sande, J. Danckaert, S. Massar, J. Dambre, B. Schrauwen, C.R. Mirasso, and I. Fischer \href{https://www.nature.com/articles/ncomms1476}{2011 \emph{Nat. Commun} \textbf{2} 468}

\bibitem{NVAR}Bollt E \href{https://doi.org/10.1063/5.0024890}{2021 \emph{Chaos} \textbf{31} 013108}

\bibitem{NG-RC} D. J. Gauthier, E. Bollt, A. Griffith, and W. A. S. Barbosa  \href{https://link.springer.com/chapter/10.1007/3-540-15644-5_26}{2021 \emph{Nat. Commun} \textbf{12} 5564}

\bibitem{dy} Jaeger H and Haas H  \href{https://www.science.org/doi/10.1126/science.1091277}{2004 \emph{Science} \textbf{304} 78}

\bibitem{reg2} Zimmermann R S and Parlitz U \href{https://aip.scitation.org/doi/full/10.1063/1.5022276}{2018 \emph{Chaos} \textbf{28} 043118}

\bibitem{Lorenz1963} E. Lorenz \href{https://link.springer.com/chapter/10.1007\%2F978-0-387-21830-4_2}{1963 \emph{J Atmos Sci} \textbf{20} 130}

\bibitem{KS} \href{https://encyclopediaofmath.org/wiki/Kuramoto-Sivashinsky_equation}{Kuramoto-Sivashinsky equation. Encyclopedia of Mathematics}

\bibitem{PRL_Large}  J. Pathak, B. Hunt, M. Girvan, Z. Lu, and E. Ott \href{https://link.aps.org/doi/10.1103/PhysRevLett.120.024102}{2018 \emph{Phys. Rev. Lett. } \textbf{120} 024102}

\end{thebibliography}
\end{document}